\documentclass[10pt,twocolumn,letterpaper]{article}

\usepackage{template_iccv/iccv}
\usepackage{times}
\usepackage{epsfig}
\usepackage{graphicx}
\usepackage{amsmath}
\usepackage{amssymb}
\usepackage{graphicx}
\usepackage{subcaption} 
\usepackage[ngerman,english]{babel}
\usepackage[format=hang]{caption}
\usepackage{placeins}            
\usepackage{paralist, tabularx}
\makeatletter
\renewcommand\tabularxcolumn[1]{>{\@minipagetrue}p{#1}}
\makeatother
\usepackage[export]{adjustbox}

\usepackage[pagebackref=true,breaklinks=true,colorlinks,bookmarks=false]{hyperref}
\iccvfinalcopy

\usepackage{listings}

\begin{document}

\definecolor{codeblue}{rgb}{0.38039216, 0.61568627, .8       }
\definecolor{codepink}{rgb}{1.        , 0.41568627, 0.83529412}
\definecolor{codegray}{rgb}{0.5,0.5,0.5}
\definecolor{codepurple}{rgb}{0.58,0,0.82}
\definecolor{backcolour}{rgb}{0.98,0.98,0.98}

\lstdefinestyle{mystyle}{
    backgroundcolor=\color{backcolour},   
    commentstyle=\color{codeblue},
    keywordstyle=\color{codepink},
    numberstyle=\tiny\color{codegray},
    stringstyle=\color{codepurple},
    basicstyle=\ttfamily\scriptsize,
    breakatwhitespace=false,         
    breaklines=true,                 
    captionpos=b,                    
    keepspaces=true,                 
    showspaces=false,                
    showstringspaces=false,
    showtabs=false,                  
    tabsize=2
}
\lstset{style=mystyle}

\lstdefinelanguage{gptprompt}{
  basicstyle=\ttfamily\footnotesize,
  numbers=left,
  keywords = {id, bbox\_extent, bbox\_center, object\_tag, caption, inferred\_query, relevant\_objects, query\_achievable, final\_relevant\_objects, explanation, query\_text},
  keywordstyle=\color{red},
  keywords = [2]{User, LLM-Planner},
  keywordstyle = [2]\color{blue},
}

\title{SceneGPT: A Language Model for 3D Scene Understanding}  
\author{
Shivam Chandhok$^{1,2}$ \\
$^1$University of British Columbia \quad $^2$Vector Institute for AI \vspace{1mm} \\
\texttt{\small \{chshivam\}@cs.ubc.ca}
}
\maketitle
\thispagestyle{empty}

\begin{abstract}
\vspace{-10pt}
Building models that can understand and reason about 3D scenes is difficult owing to the lack of data sources for 3D supervised training and large-scale training regimes. In this work we ask “How can the knowledge in a pre-trained language model be leveraged for 3D scene understanding without any 3D pre-training”.
The aim of this work is to establish whether pre-trained LLMs possess priors/knowledge required for reasoning in 3D space and how can we prompt them such that they can be used for general purpose spatial reasoning and object understanding in 3D.
To this end, we present SceneGPT, an LLM based scene understanding system which can perform 3D spatial reasoning without training or explicit 3D supervision. The key components of our framework are - 1) a 3D scene graph, that serves
as  scene representation, encoding
the objects in the scene and their spatial relationships 2) a pre-trained LLM that can be adapted with in context learning for 3D spatial reasoning. We evaluate our framework qualitatively on  object and scene understanding
tasks including object semantics, physical properties \& affordances (object-level) and spatial understanding (scene-level). 
\end{abstract}
\vspace{-0.3in}
\section{Introduction}
Developing systems that can understand the 3D world is an important goal in computer vision. Such systems require a semantically rich 3D representation that embeds objects in a spatial structure. Additionally, scene understanding systems require the ability to understand and reason about natural-language queries given a particular scene as context.
Traditional scene understanding approaches \cite{trad1,trad2,trad3,trad4} rely on specialized modules for object/spatial understanding and require 3D supervised training on a set of objects for which labelled data is curated. This creates a bottleneck due to lack of 3D data sources and infeasible large-scale training regimes. Also, these models have modules trained on datasets with limited set of classes \cite{dataset1,dataset2} for a particular specialized task (say 3D segmentation, localization) and cannot generalize to new queries, tasks and objects inhibiting their use in real-world. There is a need to develop open-vocabulary systems that can handle diverse queries (or tasks) and generalize to novel objects without 3D training.

Recent advances in large language models (LLMs) \cite{llama1,llama2,gpt} have shown their capabilities on visual understanding and general purpose reasoning. They possess knowledge about object semantics, understand free-form textual instructions and  generalize well to novel scenarios. However, it is unclear whether LLMs possess priors or knowledge required for 3D spatial reasoning.

To this end, we ask "How can the knowledge in a pre-trained large language model (LLMs) be leveraged for 3D scene understanding
without any 3D pre-training”. Specifically, we develop 
SceneGPT, a framework that leverages knowledge in LLMs for 3D scene understanding tasks.\\
Our \textbf{key contributions} are as follows -
\begin{compactitem}
\item We show that LLMs possess priors for 3D scene understanding and simple in-context prompting can unleash such capabilities without any supervision or fine-tuning.

\item We develop SceneGPT which combines an open-vocabulary scene representation with an LLM and shows promising results on diverse object and scene-level queries.
\end{compactitem}

\begin{figure*}
  \centering
  \includegraphics[width=\textwidth]{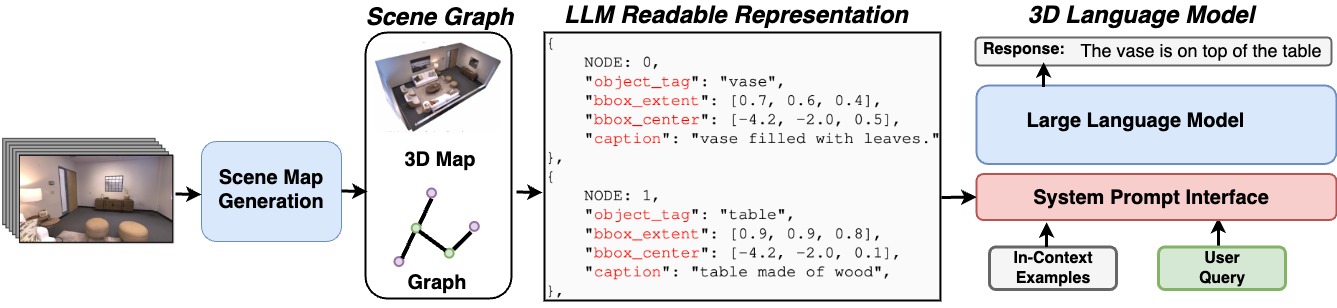} 
  \caption{Overall pipeline for our SceneGPT model. Our framework takes as input RGBD frames of a scene and generates a semantically rich 3D scene graph representation. This is converted to a LLM readable format (json) and fed to a large language model like GPT-4. The system prompt creates an interface between the scene representation and LLM and takes in the user query, scene graph json and in-context examples which guide the output of the LLM. The large language model generates a reasonable response grounded to the input scene to address the user query.}
  \label{fig:myfigure}
\end{figure*}
\section{Background and Related Work}
\paragraph{3D Scene Representations} A pivotal component of any scene understanding system is a open-vocabulary representation which embeds knowledge about object semantics in a 3D spatial structure \cite{Peng2023OpenScene,map1,map2,conceptfusion,map3,conceptgraphs}. Recent advances in multimodal foundational models has motivated methods to use priors from foundational models like CLIP for building 3D scene representations \cite{Peng2023OpenScene,map1,conceptfusion,map3}. Some notable efforts include  OpenScene \cite{Peng2023OpenScene}, which computes per-pixel features using 2D open-vocabulary segmentors
and  distills them into a networks operating over 3D data through training with CLIP features as supervision.
On the other hand, ConceptFusion \cite{conceptfusion} takes this futher to build a completely training free open-set multimodal 3D map with local and global object embeddings from CLIP. The 3D map is queryable for concepts from multiple modalities such as images, text, audio, and clicks with simple similarity operation with pixel-aligned CLIP features.  A recent approach ConceptGraphs \cite{conceptgraphs} improves over these methods to build a sparse open-vocabulary map which encodes object semantics (with the help of priors from multimodal language models like LLaVA \cite{llava}) and spatial structure as a scene graph. This approach is much more lightweight, scalable and efficient than dense feature methods and generalizes to novel objects and scenes.

For our work, we use the ConceptGraphs \cite{conceptgraphs} as the open-vocabulary scene representation which is interfaced with an LLM to develop our 3D understanding framework.
 \vspace{-5pt}
\paragraph{LLMs and prompting techniques}:
 LLMs are parameterized models trained on large corpus of free-form text from the web using language modeling objectives such as next word (or token) prediction \cite{llama1,llama2,gpt}.
 Recent models like GPT-3.5/4 \cite{gpt} (closed-source) and LLama-2 (open-source) \cite{llama2} have shown excellent capabilities on commonsense, logical understanding and visual reasoning tasks . 
 
 Given the large size and parameters of LLMs, many recent efforts have focused on developing techniques to adapt LLM for given task without updating their weights or fine-tuning \cite{weng, incontext}. This include in-context prompting (or few-shot prompting) which help steer the respnse of an LLM and align it requirements of a given task without updating weights \cite{weng, incontext}. Furthermore, techniques like prompt engineering and chain-of thought prompting \cite{chainofthought} help LLMs reason about complex task better and enhance their output quality and performance. 
 
 In this work, we use GPT-4 LLM and in-context prompting
 to develop our 3D scene understanding framework.
 \vspace{-5pt}
\paragraph{Spatial understanding with LLMs}:
Very recently there has been interest in using knowledge in LLMs for spatial trajectory prediction in videos \cite{video} and 3D reasoning tasks \cite{3dllm}. LLM grounded video diffusion \cite{video} uses few-shot prompting to predict location and layout of objects (i.e bounding boxes) in consecutive 2D frames of a video. On the other hand, 3D-LLM \cite{3dllm} uses 3D supervised joint training with a language model for scene understanding. However, unlike our approach, it requires 3D dataset for supervision and large-scale joint training regimes.
\section{Method}
\subsection{Overview}
Our goal is to use the prior knowledge in pre-trained LLMs and build a 3D scene understanding system. Figure \ref{fig:myfigure} shows the overview of our overall pipeline.
It consists of two main parts 1) Generating (llm compatible) Scene representation 2) Large Language Model.\\
On a high-level, our model takes as input RGBD frames of a scene and generates a semantically rich 3D map with object representations and graph structure. This scene representation is then converted to a LLM readable format and interfaced with a pre-trained large language model with the help of a system prompt. The LLM learns from few demonstrations of in-context examples and answers to user queries given an input scene. Let us further discuss the parts of our pipeline in detail.
\subsection{Scene Representation}
An effective 3D representation of a scene should be semantically rich, encode spatial relationships between entities and generalize to novel objects in a zero-shot way without training. In this work we use ConceptGraphs \cite{conceptgraphs} which is an an open-vocabulary graph-structured representation built by leveraging priors from 2D foundation models. This allows us to build our 3D map without any 3D data supervision or large-scale training regimes.
\paragraph{Graph Structure}: We represent scenes with an open-vocabulary 3D Scene Graph (3DSG) $\mathcal{G} = \langle \mathcal{N}, \mathcal{E} \rangle$. Each node (or vertex) $n_i \in \mathcal{N}$ in the graph corresponds to an object of interest, while the edges $e_i \in \mathcal{E}$ denote spatial relationships between pairs of objects.\\
Each node in the 3DSG additionally encodes semantic, geometric and spatial localisation information about their respective objects. Specifically, in this work, we encode \texttt{object tag} and \texttt{object caption} which name and describe the object of interest. The geometric size (\texttt{bbox extent}) and 3D position of the object (\texttt{bbox center}). And finally the \texttt{color} and \texttt{material} properties.
\vspace{-13pt}
\paragraph{LLM Readable Representation}: In order to pass our scene representation as context to LLM we first convert  our scene graph to LLM readable/compatible format (see example \textbf{Listing 1}). Motivated from previous works \cite{video,conceptgraphs}, we use the JSON data structure due to its ability to encode diverse data in key-value pair dictionary format. Furthermore, various LLMs like GPT4 (which we use in this work) are specifically trained to comprehend and reason about data in JSON format for code generation and function calling tasks. 
\vspace{-0.3in}
\paragraph{Scenegraph generation}: We follow ConceptGraphs framework with some modifications and discuss the scenegraph generation later in implementation details (Section 3).
\begin{lstlisting}[language=gptprompt, caption=Sample of our 3D Scene Graph in LM readable JSON format. Objects are represented as nodes with semantic; geometric and physical properties.]
[
    {
        "id": 0,
        "bbox_extent": [0.7,0.6,0.4],
        "bbox_center": [-4.2,-2.0,0.1],
        "object_tag": "vase",
        "caption": "The central object in this image is a vase filled with green leaves.",
        "color": "silver",
        "material": "metal/silver"
    },
    .
    .
    
    {
        "id": 5,
        "bbox_extent": [0.9,0.7,0.2],
        "bbox_center": [-4.5,-1.6,0.1],
        "object_tag": "mirror",
        "caption": "The central object in this image is a mirror.",
        "color": "brown",
        "material": "wood"
    },
    
\end{lstlisting}
\begin{figure*}[ht]
\begin{lstlisting}[language=gptprompt, caption=System prompt for creating an LLM interface with our scene. The prompt takes as input the scenegraph json; user query and in-context examples. It defines a step-by-step chain of thought structure for the LLM response to break down complex tasks and reason about them.]
"""The input to the model is a 3D scene description in a JSON format. 
Each entry in the JSON describes one object in the scene, with the following seven fields:

1. "id": a unique object identifier
2. "bbox_extent": extents of the 3D bounding box for the object. 
3. "bbox_center": centroid of the 3D bounding box for the object. 
4. "object_tag": a name for the object
5. "caption": a brief caption for the object
6. "color": the color of the object
7. "material": the material of the object

The user will ask questions, and the task is to answer various user queries about the 3D scene. 

OUTPUT FORMAT:

STEP1 - inferred_query: your interpretaion of the user query
STEP2 - relevant_objects: python list of ONLY TOP 2 most relevant object ids for the user query.
STEP3 - reason for relevance: reason for considering objects in previous steps relevant
STEP4 - Final Answer: A textual response and valid JSON with final "object_tag" and "object_id" to the original input question.
STEP-5 - Explanation: A brief explanation and reason of relevance for each relevant object and how it addresses the task. 


HERE IS YOUR INPUT:

Following is the 3D scene description : {scenegraph}
Following are the in-context examples: {examples}
Question: {input}"""
\end{lstlisting}
\end{figure*}

\subsection{LLM Interface and In-context Learning}
\vspace{-5pt}

Now that we have the LLM compatible graph representation for our 3D scene, we need to build an interface that will allow our LLM to communicate with the scene and ground its responses to the objects and structure of the given scene.
\vspace{-23pt}
\paragraph{LLM Interface} To setup a communicative interface between the scene and the LLM, we design a system prompt (taking motivation from \cite{video,conceptgraphs}) that informs the LLM about the scene representation and instructs it to take in a user query and respond in a desired format. 

Listing 2 shows the system  prompt we use to interface our scenegraph with the LLM. The prompt takes as input the scenegraph json representation, the user query and in-context demonstrations (\textbf{Lines 23-27)}. 

 First, we describe the major fields in our scene graph so the LLM understands how to interpret the scene in context to our scene understanding problem \textbf{(Listing 2, Lines 1-10)}. Next, we make it aware of the task it is expected to perform and give it an explicit output structure it should follow to reason about the task at hand and respond. 
\vspace{-10pt}
\paragraph{Chain of thought Output Structure}: On \textbf{Listing 2, Lines 14-20} of our system prompt, we use a form of chain-of-thought prompting to help LLM reason step by step about complex queries. This technique helps the LLM break a complex task into simpler sub-tasks which are sequentially performed to reach the final goal \cite{chainofthought}. In the context to our scene understanding problem, we first ask the LLM to output its interpretation of the user query and the task it has to perform \textbf{(Line 16)}. Then it retrieves the objects relevant to the given query and outputs its understanding of reason for their relevance \textbf{(Line 17-18)}. We find that forcing the LLM to explain the relevance helps it retrieve relevant objects better. Finally, we ask it to output the final answer to the query with an explanation \textbf{(Line 19-20)}. This step-by-step process helps the LLM decompose a complex task into sub-steps and better understand what the user query wants and align the output with user intent \cite{video,conceptgraphs,chainofthought}.
\begin{lstlisting}[language=gptprompt, caption= Sample of in-context example to guide our LLM to perform spatial understanding in 3D space]
QUESTION = "Where is the white couch (id: 28) with respect to pillow (id:27) "

ANSWER =[ 
    "The location of white couch (id: 28) w.r.t pillow (id:27) can be determined using the field "box_center" which determines the location of the objects. The box_center for white couch (id: 28)is [2.8, 2.3, -1.2] and pillow (id:27) is [2.9, 2.5, -0.8].
    
    The first entry corresponds to x coordinate, the second entry corresponds to y coordinate and third entry tells the height in z coordinates.

    The x-axis value in coordinate is very close i.e 2.8 for white couch and 2.9 for pillow. 
    Similarly, y-axis value in coordinate system is very close i.e 2.3 for white couch and 2.5 for pillow. This means that the objects are at the same position and close by.
    
    Since the z-axis value which determines height in coordinate system is higher i.e -0.8 for pillow than -1.2 for white couch.
    The pillow is on top of the white couch"]
    
\end{lstlisting}
\begin{lstlisting}[language=gptprompt, caption=Example of in-context demonstration to guide our LLM to perform geometric reasoning about object dimensions and sizes]
QUESTION = ["Which is bigger the white couch or the pillow (id:27)" ]

ANSWER =[ 
    "The size of white couch (id: 28) compared to the pillow (id:27) can be determined using the "box_extent" field which determines the size of the bounding box of objects.
    
    The first entry corresponds to size in x dimension, the second entry corresponds to size in y dimension and third entry tells the height in z dimension.

    The box_extent for white couch (id: 28)\
    is [1.0, 0.9, 0.6] and pillow (id:27) is [0.7, 0.6, 0.3]. Comparing these sizes of boounding boxes it is clear that white couch (id: 28) is bigger."]
\end{lstlisting}
\paragraph{In-Context prompting:}
To unleash spatial reasoning and 3D understanding capabilities, we use in-context prompting which allows us to adapt the LLM for our use case without updating weights or specialized training regimes.

Specifically, we provide LLM with few demonstrative examples with  input, output and reasoning. We incorporate multiple (i.e 2 in this work) demonstration examples across different tasks that include spatial reasoning and geometric understanding.

Listing 3 and 4 show  the in-context examples we use for spatial and geometric aspects in the scene respectively. In these examples, we help the LLM interpret the localization information (i.e \texttt{box center} field) and object dimensions (i.e \texttt{box extent} field) in 3 dimenison coordinate system (i.e x, y and z coordinate axis) and provide appropriate reasoning for input queries. This guides the LLM response later at inference where the LLM can extend the knowledge learned from these examples to a new query and task scenerio with novel objects for scene understanding.

\paragraph{Implementation Details}
We briefly describe the implementation where for scenegraph generation we follow the ConceptGraphs \cite{conceptgraphs} with off-the-shelf models RAM \cite{ram}, GroundingDINO \cite{dino} and SAM \cite{sam} model for tagging objects, generating bounding boxes for tags and prompting SAM with boxes to get object segmentations, respectively. We use LLaVA-13B \cite{llava} model as our multimodal language model to get object tags and captions from crops. We go a step further and also ask LLaVA to predict the color and material of the objects for a more informative scene representation. Finally, we use the gpt4-16k model with 16k token limit as our large language model.
\section{Results}
\begin{figure*}
  \centering

  \begin{subfigure}{0.49\textwidth}
  \hspace{-45pt}
    \includegraphics[width=1.2\linewidth]{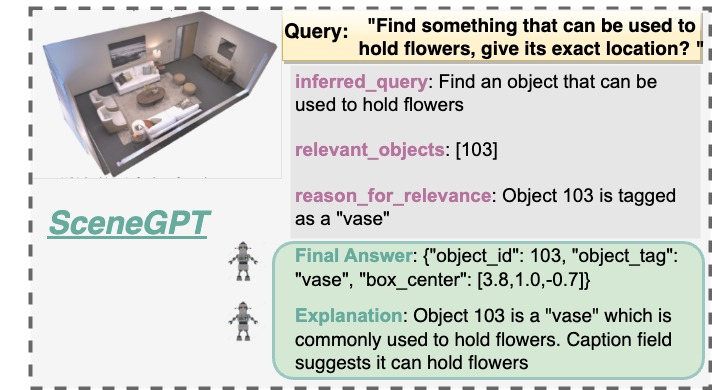} 
    \caption{Affordance based object understanding}
  \end{subfigure}
  \hfill
  \begin{subfigure}{0.49\textwidth}
  \hspace{-5pt}
    \includegraphics[width=1.2\linewidth]{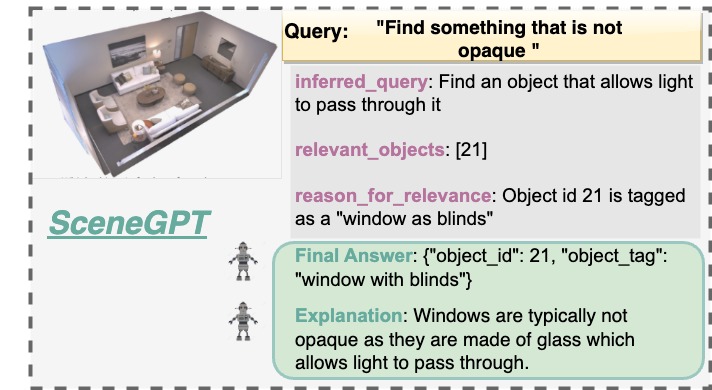} 
    \caption{Negation based object understanding}
  \end{subfigure}

  \begin{subfigure}{0.49\textwidth}
  \hspace{-45pt}
    \includegraphics[width=1.2\linewidth]{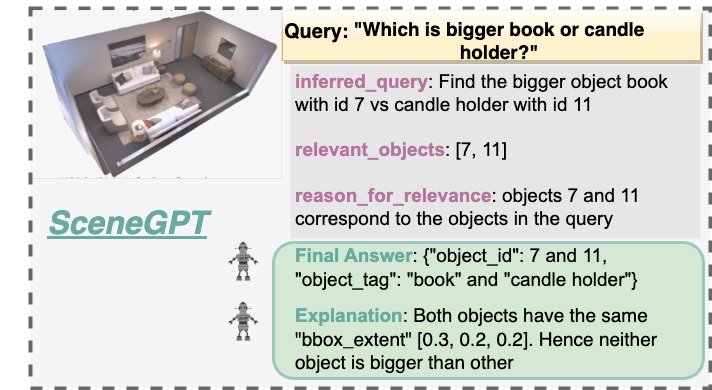} 
    \caption{Geometric understanding results}
  \end{subfigure}
  \hfill
  \begin{subfigure}{0.49\textwidth}
  \hspace{-5pt}
    \includegraphics[width=1.2\linewidth]{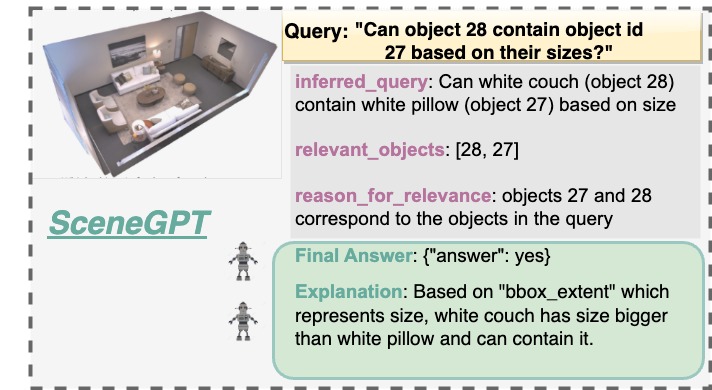} 
    \caption{Geometric reasoning results}
  \end{subfigure}
  \begin{subfigure}{0.49\textwidth}
  \hspace{-45pt}
    \includegraphics[width=1.2\linewidth]{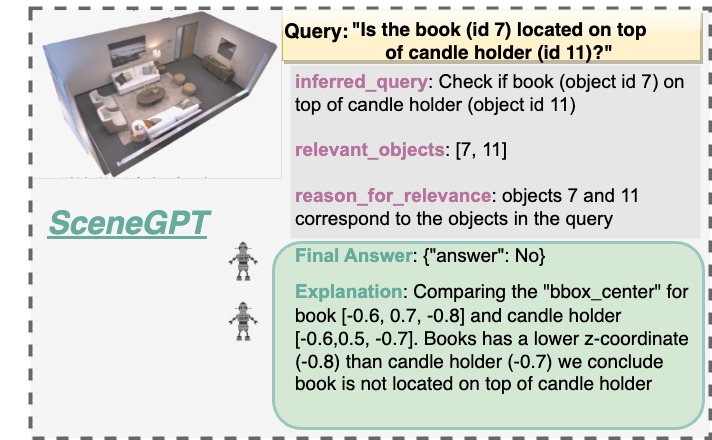} 
    \caption{Spatial Reasoning results}
  \end{subfigure}
  \hfill
  \begin{subfigure}{0.49\textwidth}
  \hspace{-5pt}
    \includegraphics[width=1.2\linewidth]{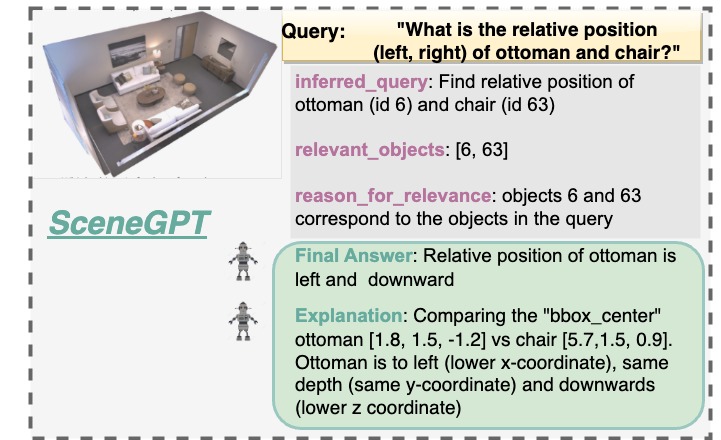} 
    \caption{Spatial Understanding results}
  \end{subfigure}

  \caption{Qualitative results for our SceneGPT 3D understanding system on object understanding queries(top row), \\geometric understanding queries (middle), and spatial understanding queries (bottom)}
  \label{fig:fullpagefigures}
\end{figure*}

In this section we discuss qualitative results for our LLM based 3D scene understanding framework. In order to validate the efficacy of our framework we show results on diverse queries which can broadly be divided into 3 groups 1) Object understanding 2) Geometric queries 2) Spatial queries. 
\vspace{-10pt}
\paragraph{Object understanding}: In this group we test our models object understanding capabilities with \textit{affordance} based and \textit{negation} queries motivated from \cite{conceptgraphs}. Figure \ref{fig:fullpagefigures} (top row) shows the qualitative results for these queries.

We first test our model on  affordance based query, where we ask our system to retrieve an object that can be used for a given task (i.e \textit{Something that can be used to hold flowers}) \textbf{Figure 2(a)}. 
Notice that our framework reasons about the user query step by step, uses its prior knowledge about affordances and grounds its response to the scene, correctly retrieving a "vase" and its exact location with the help of \texttt{box center}. Next, we test our method on a negation query where we ask it to retrieve an object that does not satisfy a particular physical property i.e (\textit{Something that is not opaque }) \textbf{Figure 2(b)}. Our model reasons about object properties and understands the negation query and finally retrieves a window which is usually transparent. These results are in alignment with some of the observations in previous works \cite{conceptfusion, conceptgraphs} which also show LLMs have object understanding capabilities.
\vspace{-10pt}
\paragraph{Geometric understanding}: In this query pool, we test the understanding of our model about object geometry and sizes. We ask it to compare the size of objects in the scene and ask it to reason about whether a given object can contain another object (\textbf{Figure 2(c) and 2(d)}). 

Thanks to our in-context examples which helped the LLM understand the relation between \texttt{box extent} and object size, we notice that the LLM accurately compares query objects (i.e size of "book" and "candle holder" \textbf{Figure 2(c)}) based on \texttt{box extent} and correctly answers the user queries. It also successfully reason for the fact that the "white couch" can contain the "white pillow" since it is larger in size and can accomodate the "white pillow" in the scene \textbf{Figure 2(d)}.
\vspace{-10pt}
\paragraph{Spatial understanding}:
 We test how well the SceneGPT model understands object orientation and relative placement of objects in 3D space. First, we test the spatial reasoning capability of our model by asking it if - \textit{The book is located on top of the candle holder} (\textbf{Figure 2(e)}). Notice that the LLM correctly views the problem separately in x, y and z dimension and reaches the conclusion that the book is not on top of the candle holder by comparing the z-coordinates. It is worth noting that the LLM has learned spatial reasoning in  3D coordinate system from the in-context examples and  seamlessly applied it to a new set of objects in a different problem setting and query. Furthermore, we take a step ahead and ask the LLM to give the spatial relationship and orientation of different objects w.r.t each other by asking \textit{the relative position of ottoman and chair} (\textbf{Figure 2(f)}).Our SceneGPT framework reasons about the location of objects in the 3 dimensions and outputs the correct spatial relationship among query objects.

\section{Failure Cases and Limitations}
The SceneGPT framework comprises of a scene graph representation (i.e ConceptGraphs in this work) and a large langugae model (i.e GPT-4 in our case) and inherits some of their fundamental limitations which we discuss in this section.
\begin{compactitem}
\item The SceneGPT framework inputs the scenegraph as a json and is limited by the context-length of the language model. We tried to run our framework on \textbf{ScanNet scenes} \cite{dataset1} and evaluate on ScanQA dataset but these scenes had more than 120 object nodes in the scenegraph which could not be taken as input by our model due to the limited context length of publically available GPT4 model.
\item The ConceptGraphs framework uses pre-trained multimodal language models (i.e LLaVA \cite{llava}) to assign object nodes, tags and captions. The multimodal LMs are close to 70\% accurate and label the objects nodes incorrectly 30 percent of the time which leads to incorrect object nodes in the scenegraph (since these models have not been fine-tuned). This limits the accuracy of our framework since some detected objects are incorrect. However, as stronger multimodal LMs are developed we hope that the object nodes predictions in the scene will become more accurate.
\end{compactitem}
\vspace{-0.1in}
\section{Conclusion and Future Work}
In this work we developed an LLM based 3D scene understanding system without 3D supervised data and large-scale training regimes. We demonstrated that a simple technique like in-context prompting can help adapt LLMs and uncover spatial reasoning capabilities in them.

In future, we will try to extend this framework to more complex tasks like navigation, trajectory prediction and task planning and validate if simple techniques like in-context learning are enough for such complex tasks.

{\small
\bibliographystyle{template_iccv/ieee_fullname}
\bibliography{main}
}

\end{document}